\documentclass[10pt,twocolumn,letterpaper]{article}

\usepackage{cvpr}              

\usepackage[dvipsnames]{xcolor}

\usepackage{caption}
\usepackage{textcomp}
\usepackage{color, colortbl}
\usepackage{mathtools}
\usepackage{appendix}
\usepackage{multirow}

\usepackage[accsupp]{axessibility} 

\definecolor{cvprblue}{rgb}{0.21,0.49,0.74}
\usepackage[pagebackref,breaklinks,colorlinks,citecolor=cvprblue]{hyperref}

\def\paperID{14900} 
\def\confName{CVPR}
\def\confYear{2024}

\title{OVFoodSeg: Elevating Open-Vocabulary Food Image Segmentation via Image-Informed Textual Representation}

\author{
 Xiongwei Wu \quad Sicheng Yu \quad Ee-Peng Lim \quad
 Chong-Wah Ngo \\[2mm]
 Singapore Management University\\[2mm]
 {\tt\small \{xwwu.2015, scyu.2018\}@phdcs.smu.edu.sg}, 
 {\tt\small\{eplim, cwngo\}@smu.edu.sg}
 \vspace{-4mm} 
}

\begin{document}
\maketitle
\begin{abstract}
In the realm of food computing, segmenting ingredients from images poses substantial challenges due to the large intra-class variance among the same ingredients, the emergence of new ingredients, and the high annotation costs associated with large food segmentation datasets.
Existing approaches primarily utilize a closed-vocabulary and static text embeddings setting. These methods often fall short in effectively handling the ingredients, particularly new and diverse ones. In response to these limitations, we introduce OVFoodSeg, a framework that adopts an open-vocabulary setting and enhances text embeddings with visual context.
By integrating vision-language models (VLMs), our approach enriches text embedding with image-specific information through two innovative modules, \eg, an image-to-text learner FoodLearner and an Image-Informed Text Encoder.
The training process of OVFoodSeg is divided into two stages: the pre-training of FoodLearner and the subsequent learning phase for segmentation. The pre-training phase equips FoodLearner with the capability to align visual information with corresponding textual representations that are specifically related to food, while the second phase adapts both the FoodLearner and the Image-Informed Text Encoder for the segmentation task.
By addressing the deficiencies of previous models, OVFoodSeg demonstrates a significant improvement, achieving an 4.9\% increase in mean Intersection over Union (mIoU) on the FoodSeg103 dataset, setting a new milestone for food image segmentation.
\end{abstract}    
\section{Introduction}
\label{sec:intro}
Food computing has garnered significant attention as a research focus due to its widespread application~\cite{David-DH-Nature2014,Boswell-FC-PNAS2018,Meyers-Im2Calories-ICCV2015,Quin-Nutrition5k-CVPR2021}: it integrates computer science with the analysis of food and dietary patterns that have a direct impact on people's healthy eating habits~\cite{min2019survey}. At the core of food computing is the task of food image segmentation~\cite{wu2021large,uecfoodpixcomplete,uecfoodpix}. It aims to identify the ingredients in a food image and locate them with pixel-level masks. This task can be easily found in many practical applications, \eg, analysis of caloric and nutritional content of food intake~\cite{min2021large,MIN2022100484}, hence, making it very important in food computing research.

\begin{figure}[tp]
\begin{center}
\includegraphics[width=1.\linewidth]{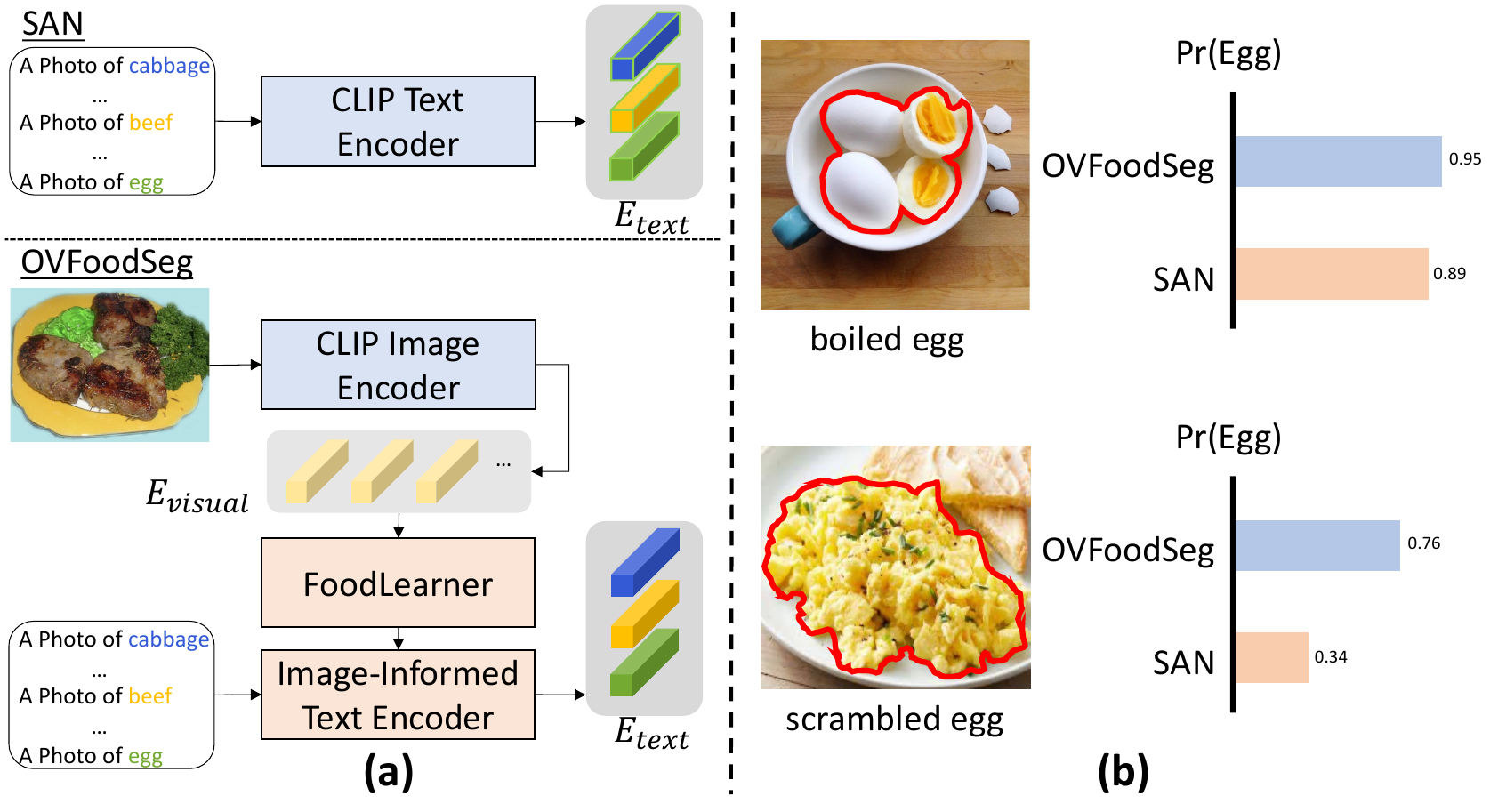}
\end{center}
\vspace{-0.15cm}
\caption{(a) Top: Conventional open-vocabulary segmentation framework Side Adaptive Network (SAN)~\cite{xu2023san}, predicts mask category logits by raw text embeddings from CLIP; Bottom: Proposed OVFoodSeg, constructs image-informed text embeddings through FoodLearner and the Image-Informed Text Encoder for mask classification.  (b) This example illustrates the use of both SAN and the proposed OVFoodSeg in identifying egg masks cooked using different methods.
}
\vspace{-0.5cm}
\label{fig:overview}
\end{figure}

The standard food image segmentation follows a pipeline~\cite{wu2021large} that begins with the annotation of a large image dataset with pixel-wise masks followed by the training of image segmentation models~\cite{SETR,chen2017deeplab}.
However, the pixel mask annotation of images is not only time consuming but also complex. 
The complexity is further exacerbated in the domain of food
as the ingredients can look very small in food images, and they may interact with one another to form obscuring boundaries.
While publicly available annotated image datasets for semantic segmentation exist~\cite{uecfoodpix,Meyers-Im2Calories-ICCV2015}, there is only one ingredient-level food segmentation dataset~\cite{wu2021large} that provides the human annotated masks of 103 ingredient labels across 7,000 images.
Even such a small dataset took one year to construct. The set of 103 ingredient labels is also too small to cover all the ingredients that are used in today's food.
To tackle the limited ingredient coverage among annotated food image datasets, we propose to address food image segmentation in the open-vocabulary setting. Under this setting, our goal is to learn a segmentation model that adapts to novel ingredients which are not found in the training data without compromising much accuracy performance.

Vision-language models (VLMs) which aligns both vision and language modalities demonstrate remarkable capabilities as a feature extractor in zero-shot classification~\cite{radford2021learning} and open-vocabulary detection tasks~\cite{gu2022openvocabulary,kuo2023openvocabulary,du2022learning}.  
One such example VLMs is CLIP~\cite{radford2021learning} which learns to encode both image and text embeddings via contrastive loss.
For image segmentation, CLIP first generates the text embeddings from the ingredient text describing the food image (top of Figure \ref{fig:overview}(a)) and then contrast with the corresponding image representation. 
While such methods achieve success with general domain images, it fails to produce similar gains due to intra-class variance of food ingredients: 
the same ingredient with different cooking methods may look very different visually, \eg, the ``egg'' class shows considerable variations in appearance between boiled egg and scrambled egg (see Figure \ref{fig:overview}(b)).
The intra-class variance is further exacerbated by unseen classes, where the model lacks prior exposure to these ingredients and the many ways they are prepared and cooked.

To address the aforementioned challenges, we introduce an Open-Vocabulary Food Segmentation framework, named \textbf{OVFoodSeg}. The framework effectively integrates the capabilities of CLIP with the image-to-text \emph{FoodLearner}, and replaces CLIP's original fixed text encoder with the proposed \emph{Image-Informed Text Encoder} (see bottom of Figure~\ref{fig:overview}(a)).
By harnessing cross-modality capabilities of CLIP, OVFoodSeg effectively transfers knowledge from seen ingredients to novel ingredients.
FoodLearner is proposed to mitigate the high intra-class variance for ingredients. Specifically, FoodLearner is a BERT-style text-to-image learner motivated by BLIP2~\cite{li2023blip}, which is designed to automatically extract visual knowledge from the food images. The extracted visual knowledge is subsequently fed into the {Image-Informed Text Encoder}, enhancing CLIP's static text representation with image-specific information. This process enables tailored adjustments to the text embeddings, effectively dealing with ingredient classes that exhibit diverse visual appearances. 
Figure \ref{fig:overview} (b) showcases an example of classifying egg prepared in different ways using both the conventional static text embedding and the proposed image-informed text embedding. In more common scenarios (such as boiled egg), both models yield reliably accurate predictions. However, in more ambiguous cases (like scrambled egg), the performance of the conventional static text embedding is poor, while the image-informed text embedding still achieves impressive results.

The training process of OVFoodSeg comprises two stages: the FoodLearner Pre-training stage and the Segmentation Learning stage. In the first stage, the FoodLearner module is trained to align visual information with textual representations. This training utilizes a large-scale dataset of food-related image-text pairs, intended to introduce the FoodLearner to a broad range of food items. Following this, the Segmentation Learning stage involves fine-tuning the FoodLearner for the open-vocabulary food image segmentation task using a specialized segmentation dataset. This stage further hones the model's ability to accurately segment food items in diverse scenarios.

To validate the effectiveness of OVFoodSeg, we conduct experiments under open-vocabulary setting based on the existing publicly available food image segmentation dataset, FoodSeg103~\cite{wu2021large}. We specifically design two open-vocabulary scenarios for the experiments: (i) a random splitting of FoodSeg103, allocating 80\% of the ingredient classes as base classes for training and the remaining as novel classes for inference, and (ii) the expansion of FoodSeg103 by introducing 92 new classes to create FoodSeg195, followed by a similar random split. The results of these experiments indicate that our OVFoodSeg outperforms the state-of-the-art open-vocabulary segmentation method, \ie, SAN~\cite{xu2023san}, by 4.9\% in mean Intersection over Union (mIoU) on novel classes in FoodSeg103 and by 3.5\% in FoodSeg195. Here is a summary of our contributions:
\begin{itemize}
    \item We introduced OVFoodSeg, an innovative framework specifically designed for open-vocabulary food image segmentation. Our OVFoodSeg employs a two-stage training and is the first to address the challenge of segmenting food ingredients at an open-vocabulary level.
    \item To tackle the issue of large intra-class variance in the visual representation of food ingredients, OVFoodSeg integrates an image-to-text learning module, FoodLearner, along with the Image-Informed Text Encoder. This integration enriches textual representations with visual knowledge derived from food images.
    \item OVFoodSeg achieves state-of-the-art performance in two open-vocabulary benchmarks for food image segmentation, demonstrating its effectiveness and advancement over existing methods.
\end{itemize}

\section{Related Work}
\label{sec:relatedwork}
\noindent{\textbf{Food-related Segmention Dataset: }}
Food image segmentation is a core problem in food computing, and constructing a large-scale datasets with pixel-wise mask annotation is the fundamentation to address this problem. 
Myers~\etal.~\cite{Meyers-Im2Calories-ICCV2015} construct Food201, a dataset containing roughly 12,000 images across 201 dish classes. Later UECFoodPix~\cite{uecfoodpix} and UECFoodPixComplete~\cite{uecfoodpixcomplete} are proposed which contain about 10,000 images across 102 dish types for segmentation. However, the annotations of these datasets are all restricted to dish-wise masks, rendering them unsuitable for ingredient segmentation. In response to this limitation, FoodSeg103~\cite{wu2021large} is created as the first food image segmentation dataset with ingredient-level annotations which is still far from enough to present the diversity of food. 
Thus we propose to address food image segmentation in the open-vocabulary setting.

\noindent{\textbf{Food Image Segmentation: }}
Concurrently, the exploration of food image segmentation frameworks is advancing. Wu~\etal~\cite{wu2021large} introduce ReLeM, a method designed to mitigate the large intra-class variance by incorporating recipe information into the visual representation. 
Wang~\etal~\cite{wang2022swin} develope a Swin Transformer-based segmenter called STPPN, which harnesses contextual information from various regions within the food image, thus enriching the global representation. 
Jaswanthi~\etal~\cite{jaswanthi2022hybrid} utilize a hybrid approach for food image segmentation, initially applying a GAN model~\cite{Isola_2017_CVPR} to generate proposal masks for food images, which are then categorized by a CNN-based recognition model. 
Lan~\etal~\cite{lan2023foodsam} introduce FoodSAM, combining the pre-trained SAM model~\cite{kirillov2023segment} with a segmentation model~\cite{SETR} trained on FoodSeg103 for high-quality mask generation.
However, these existing algorithms in food image segmentation predominantly focus on close-set learning, limiting their adaptability to a wider array of ingredient categories not included in the training set.

\noindent{\textbf{Open-Vocabulary Segmentation:}}
Open-vocabulary segmentation aims to identify objects with pixel-wise masks beyond the classes seen during training. Early work~\cite{bucher2019zero,xian2019semantic} concentrated on creating joint embeddings that link image pixels with class concepts. Recently, CLIP-based frameworks have made significant strides in this field. Zhou~\etal~\cite{zhou2022extract} present MaskCLIP, which uses a frozen CLIP model to generate pseudo pixel labels for segmentation model training. OpenSeg~\cite{ghiasi2022scaling} enhances segment-level visual features with text embeddings through region-word associations. Xu~\etal.~\cite{xu2022simple} develop SimSeg, which produces class-agnostic masks with a mask generator, followed by classification with a CLIP-based classifier. Building upon this, SAN~\cite{xu2023san} employs CLIP directly for mask generation, foregoing the complex mask generator. Liang~\etal~\cite{liang2023open} introduce OVSeg, which fine-tunes the original CLIP image encoder with masked images to improve segmentation understanding. These approaches tend to use static text embeddings from CLIP encoded class names, neglecting the variability of image content, which can be detrimental when handling images with large intra-class variance. Wang~\etal~\cite{wang2023hierarchical} propose HIPIE, which integrates visual features into text embeddings through attention modules. However, the simple attention modules struggle to handle the misalignment between the visual and text spaces, while our proposed FoodLearner is capable of aligning the visual and text spaces with much less computational cost.

\section{OVFoodSeg}
To address the significant challenge of large intra-class variance in ingredients for effective food image segmentation in open-vocabulary settings, we present OVFoodSeg. Inspired by the recent success of Vision-Language Models (VLMs) in open-vocabulary settings for various tasks, OVFoodSeg is developed on the foundation of a frozen CLIP model. CLIP integrates both image encoder and text encoder to embed the two modalities. To mitigate the large intra-class variance, OVFoodSeg incorporates two critical components, namely the \textit{FoodLearner} and \textit{Image-Informed Text Encoder}.  FoodLearner is responsible for extracting visual information from food images, which is then utilized by the Image-Informed Text Encoder to enhance CLIP's text embeddings.
The training procedure of OVFoodSeg is structured into two stages: first stage to pre-train the FoodLearner with a large-scale dataset of food-related image-text pairs, and the second stage to fine-tune the pre-trained FoodLearner to perform segmentation task. In this section, we will introduce two training stages along with corresponding modules.

\begin{figure*}[ht]
\begin{center}
\includegraphics[width=1.0\linewidth]{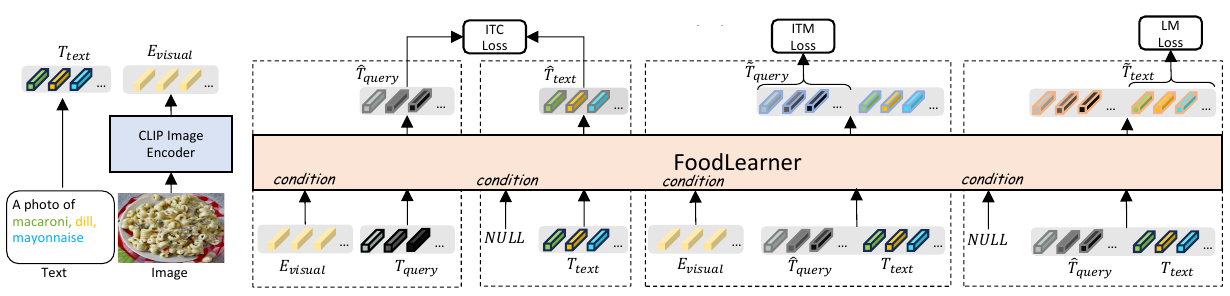}
\end{center}
\vspace{-0.13cm}
\caption{The figure depicts the pipeline of Stage I FoodLearner Pre-training. Stage I is dedicated to pre-training the FoodLearner module with image-text pairs pertinent to food so that the visual information closely related to the accompanying text will be extracted to enrich the text representation.
}
\label{fig:pipeline_pre}
\end{figure*}

\begin{figure*}[ht]
\begin{center}
\includegraphics[width=0.9\linewidth]{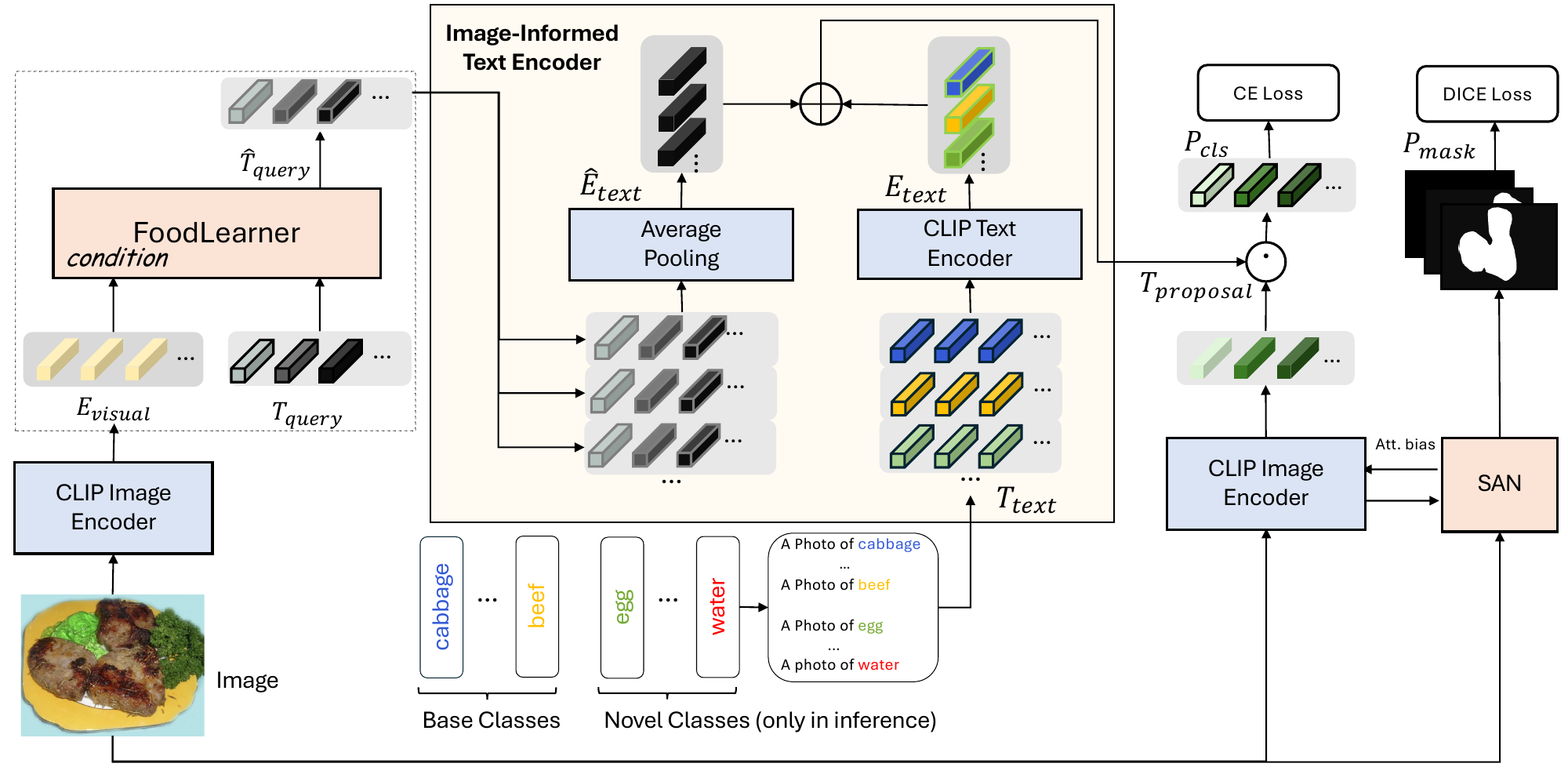}
\end{center}
\vspace{-0.13cm}
\caption{This figure depicts the pipeline of Stage II Segmentation Learning, focusing on training the segmenter using image-informed text embeddings. The FoodLearner extracts image-specific information which are then combined with the text embeddings from Image-Informed Text Encoder to produce the final image-informed text embeddings. Noted that modules with the same name share the parameters, \ie, CLIP image encoder and CLIP text encoder.
}
\vspace{-0.3cm}
\label{fig:pipeline_seg}
\end{figure*}

\subsection{Stage I Training: FoodLearner Pre-training}
Figure \ref{fig:pipeline_pre} presents the training block for Stage I, focusing on the pre-training of the FoodLearner using pairs of images and texts related to food. The objective is to explicitly pre-train the FoodLearner so that the visual information closely related to the accompanying text will be extracted to enrich the text representation. To achieve this, we simultaneously optimize three types of loss: Image-Text Contrastive loss (ITC Loss), Image-Text Matching loss (ITM Loss), and Language Modeling loss (LM Loss). These losses function in concert, sharing the same model parameters throughout the pre-training process. In the following, we outline the FoodLearner structure and three objective losses utilized during pre-training.

\noindent\textbf{FoodLearner} is implemented with the transformer structure~\cite{devlin2018bert}, \ie, multiple transformer blocks, each featuring a self-attention layer, a cross-attention layer, and a feed-forward network. 
A collection of learnable query tokens $T_{query}$ is introduced to interface with the conditional visual embedding through the cross-attention layer to capture visual knowledge.
The primary input of FoodLearner is a sequence of tokens, where these tokens can be either learnable query tokens, text tokens, or a combined sequence of query and text tokens. Additionally, FoodLearner may also receive a conditional input, \ie, visual embedding $E_{visual}$, from the CLIP image encoder. 
Thus the output of FoodLearner, \ie, FoodLearner enriched token, could be denoted as $\mathrm{FL}(E_{visual}, T_{query})$ or $\mathrm{FL}(NULL, T_{text})$, where the former is a conditional input.
Noted that the cross-attention layer is activated only when the conditional input is not $NULL$.

\noindent{\textbf{Image-Text Contrastive Loss}} 
is designed to promote a higher similarity for positive image-text pairs as opposed to negative pairs. 
We first generate the FoodLearner enriched tokens for both image and corresponding prompted text, \ie, enriched query tokens $\hat{T}_{query}=\mathrm{FL}(E_{visual}, T_{query})$ and enriched text tokens $\hat{T}_{text}=\mathrm{FL}(NULL, T_{text})$.
Following this, we calculate the pairwise similarity between each enriched query token and enriched text token and select the highest value to represent the image-text similarity. For each image and text, we compute the softmax-normalized image-to-text and text-to-image similarities $p^{t2i}_i$ and $p^{i2t}_i$ as:
\begin{equation}
    p^{t2i}_i = \frac{max_{t}( \phi *\left \langle \left [  \hat{T}^t_{{query}} \right ]_i, \left [  \hat{T}_{{text}}^{} \right ]_i  \right \rangle)}{\sum_{j}max_{t}( \phi *\left \langle \left [  \hat{T}^t_{{query}} \right ]_j, \left [  \hat{T}_{{text}}^{} \right ]_i  \right \rangle)}
\end{equation}
\begin{equation}
    p^{i2t}_i = \frac{max_{t}( \phi *\left \langle \left [  \hat{T}^t_{{query}} \right ]_i, \left [  \hat{T}_{{text}}^{} \right ]_i  \right \rangle)}{\sum_{j}max_{t}( \phi *\left \langle \left [  \hat{T}^t_{{query}} \right ]_i, \left [  \hat{T}_{{text}}^{} \right ]_j  \right \rangle)}
\end{equation}
Here $\left [  \hat{T}_{{query}} \right ]_i$ and $\left [  \hat{T}_{{text}}^{} \right ]_i$ correspond to the $i$-th image's enriched query tokens and enriched text tokens, respectively. The notation $\langle \ ,\ \rangle$ computes the cosine similarity. The variable $t$ indicates the index of the query token exhibiting the highest similarity to the text embedding contrasted, and $\phi$ is the temperature parameter used to scale the similarity (set to 10 in this paper). The ITC Loss is then computed by the Cross-Entropy loss (CE) as:
\begin{equation}
    L_{ITC} = \frac{1}{2}*(\mathrm{CE}(p^{t2i}, GT_{t2i}) + \mathrm{CE}(p^{i2t}, GT_{i2t}))
\end{equation}
where $GT_{t2i}$ and $GT_{i2t}$ indicate the ground truth labels, with positive pairs assigned label 1 and negative pairs label 0.

\noindent{\textbf{Image-Text Matching Loss}} is designed to learn fine-grained alignment between image and the corresponding prompted text. A binary classifier is learned here to predict whether the image-text pair is matching or not. The process begins by concatenating 
$\hat{T}_{{query}}$ and  $T_{{text}}$ and feeding them with visual embedding into the FoodLearner, \ie, $\Tilde{T}_{query}=\mathrm{FL}(E_{visual}, \hat{T}_{{query}};T_{{text}})$, where $;$ denotes concatenation. 
Different from ITC, $\hat{T}_{{query}}$ and $T_{{text}}$ could interact within the self-attention layer, thus enabling the query tokens assimilate multimodal information from the text. Subsequently, the resulting output query tokens $\tilde{T}_{query}$ are passed through a fully-connected layer to calculate the match probability. Given that there are $Q$ query tokens, the final probability value $p^{itm}$ is derived by averaging these values.:
\begin{equation}
    p^{itm} = \frac{1}{Q}\sum_{t=1}^{Q} \mathrm{FC}( \left [  \Tilde{T}^t_{{query}} \right ])
\end{equation}
The ITM Loss is then computed as:
\begin{equation}
    L_{ITM} = \mathrm{CE}(p^{itm}, GT_{itm})
\end{equation}
where the $GT_{itm}$ indicates the ground truth labels.

\noindent{\textbf{Language Modeling Loss}} 
aims to facilitate the generation of text informed by image content. The process also begins by concatenating the query tokens, $\hat{T}_{{query}}$ with the text tokens $T_{{text}}$, which are then passed into FoodLearner to generate $\Tilde{T}_{text}=\mathrm{FL}(NULL, \hat{T}_{{query}};T_{{text}})$. Within FoodLearner module, 
we try to reconstruct original input words given the query tokens.
Thus we optimize the auto-regressive LM Loss $L_{LM}$ on the $W$ text tokens to maximize the probability of the reconstructed text:
\begin{equation}
    p^{word}_w=\mathrm{FC}(\Tilde{T}_{text}^w)
\end{equation}
\begin{equation}
    L_{LM} = \frac{1}{W}\sum_w\mathrm{CE}(p^{word}_w,GT_{word})
\end{equation}

Since $T_{{text}}$ interacts exclusively with $\hat{T}_{{query}}$ and not directly with $E_{visual}$, $\hat{T}_{{query}}$ are fine-tuned to encapsulate the visual information most pertinent to the text.

Finally, the loss function of Stage I in training FoodLearner is:
\begin{equation}
    L_{{Stage I}} = L_{ITC} + L_{ITM} + L_{LM}
\end{equation}

\subsection{Stage II Training: Segmentation Learning}
Figure~\ref{fig:pipeline_seg} illustrates the training process of Stage II for OVFoodSeg, focusing on the learning of the food image segmentation framework. During this stage, we utilize the FoodLearner, which was pre-trained in Stage I, to extract image-specific information. This information, in conjunction with the text embeddings of the target ingredient classes, is then fed into the Image-Informed Text Encoder to generate image-informed text embeddings, which are subsequently used for the segmentation learning.

Firstly, the FoodLearner utilizes learnable query tokens $T_{query}$ to extract image-specific information from the visual embeddings $E_{visual}$ of the input images: $\hat{T}_{query}=\mathrm{FL}(E_{visual}, T_{query})$. We calculate the average of $\hat{T}_{query}$ across $Q$ query tokens as the visual representation $\hat{E}_{{text}}$:
\begin{equation}
    \hat{E}_{{text}} = \text{Average-Pooling}({T}_{{query}}) 
\end{equation}
Then we individually pair the name of each ingredient class with a pre-designed template (using ``A Photo of \{\}'' in the paper) to create the text token for that specific ingredient class. The text tokens $T_{{text}}$, each each comprising $L$ vectors where $L$ denotes the vocabulary size. The text embeddings for the original text tokens are encoded as:
\begin{equation}
    {E}_{{text}} = \text{CLIP-TE}({{T}}_{{text}}) 
\end{equation}
We create the image-informed text embeddings by integrating $\hat{E}_{{text}}$ with ${E}_{{text}}$ using an element-wise summation operation:
\begin{equation}
    \tilde{E}_{{text}} = \hat{{E}}_{{text}} + {{E}}_{{text}}
\end{equation}
$\tilde{E}_{{text}}$ are then input into a standard open-vocabulary segmentation framework to facilitate segmentation learning. In this paper, we follow the pipeline of SAN~\cite{xu2023san}, which is fine-tuned with two objective losses: Cross-Entropy loss for classification and Dice loss~\cite{sudre2017generalised} for mask prediction. For classification, the CLIP image encoder generates a set of proposal tokens ${T}_{{proposal}}$, each conditioned on the attention bias from the SAN module and corresponding to a specific region of the input image. Subsequently, the image-informed text embeddings $\tilde{E}_{{text}}$ are  engaged in a dot product with ${T}_{{proposal}}$, yield the class probability distributions:

\begin{equation}
    P_{cls}^{i} = \frac{\text{exp}(\tau*<{T}_{{proposal}}, \tilde{E}^i_{{text}}>)}{\sum_j \text{exp}(\tau*<{T}_{{proposal}}, \tilde{E}^j_{{text}}>)}
\end{equation}
Here $P_{cls}^i$ represents the predicted probability values for the $i$-th class, and $\tau$ is the temperature used to re-scale the similarity (set 100 here). For mask prediction, the framework incorporates SAN module, a lightweight Vision Transformer, that calculates binary masks aligned with ${T}_{{proposal}}$ from the classification branch.
Finally, the classification branch is optimized by Cross-Entropy loss (CE) while the mask prediction module is optimized by Dice loss (DICE:
\begin{equation}
    L_{{Stage II}} = \text{CE}(P_{cls}, GT_{cls}) + \text{Dice}(P_{mask}, GT_{mask}) 
\end{equation}
Here $GT_{cls}$ and $GT_{mask}$ represent the ground truth for class and mask, respectively. During the Stage II training, we maintain the parameters of the CLIP text encoders fixed and focus on training the FoodLearner and the SAN exclusively.

\section{Experiment}
In this section, we first detail our experimental setup, including dataset configuration and model implementation. We then present the performance of our model on two open-vocabulary food image segmentation benchmarks. Lastly, we provide an ablation study to evaluate the impact of different components and settings within OVFoodSeg based on FoodSeg103 dataset followed by the case study.

\subsection{Experimental Setup}~\label{sec:exp}
\noindent{\textbf{Training Dataset of Stage I}}: 
For Stage I pre-training, we utilized the Recipe-1M+ dataset~\cite{marin2018recipe1m+}, which is currently the most extensive dataset of image-recipe pairs available, containing approximately 1 million culinary recipes. Each recipe is detailed with the dish's name, cooking methods, an ingredient list, and about 10 images representing the dish from various restaurants. 
This dataset, with its extensive collection of food photographs and ingredient lists, is particularly well-suited for Stage I training. However, it's important to note that the ingredient lists in the dataset often include ``invisible'' ingredients such as sugar, oil, and salt. These ingredients, while integral to the recipes, contribute to data noise since they are invisible in the images.
To refine the quality of our ingredient data, we utilize ChatGPT to obtain the most visually evident ingredients for each recipe\footnote{We request ChatGPT prompted with ``Given a recipe name as \{\}, generate a short description. Only mention the main ingredients that can be seen in the dish (no more than 5).''}.
We then leveraged the cleaned ingredient lists as text information, paired with the corresponding images, to pre-train the FoodLearner module.

\noindent{\textbf{Training Dataset of Stage II}}: 
For Stage II segmentation learning, we conduct experiments on two benchmarks: FoodSeg103 and FoodSeg195. FoodSeg103 comprises approximately 7,000 images across 103 ingredient classes. We randomly select 20 classes as novel classes and use the remaining 83 as base classes. Expanding upon FoodSeg103, we add an additional 92 classes to create the larger dataset FoodSeg195\footnote{The images were collected from the residents of a country for food logging through a government agency. Due to the confidentiality agreements, FoodSeg195 cannot be made publicaly available at the moment.}, which includes around 18k images for training and 16k for testing, totaling 113k annotated masks. From FoodSeg195, we randomly designate 40 classes as novel, with the rest serving as training data. The annotations of novel classes are blocked during training, and the test sets of FoodSeg103 and FoodSeg195 are used for evaluating the model performance. To mitigate the impact of randomness on experimental results, we conducted three random class splits for both the FoodSeg103 and FoodSeg195 datasets in the main experiments. These experiments were carried out over three iterations, and we reported the average values and standard deviations. In the ablation study, we report results from only the first split based on one experiment. For the details of the class splits, please refer to the appendix.

\begin{table*}[htp]
\resizebox{1.0\textwidth}{!}{
\begin{tabular}{|l|l>{\color{gray}}l>{\color{gray}}l|l>{\color{gray}}l>{\color{gray}}l|l>{\color{gray}}l>{\color{gray}}l|}
\hline
\multicolumn{1}{|l|}{\multirow{2}{*}{Method}} & \multicolumn{3}{c|}{Split 1}                                                          & \multicolumn{3}{c|}{Split 2}                                                          & \multicolumn{3}{c|}{Split 3}                                                          \\
\multicolumn{1}{|c|}{}                        & \multicolumn{1}{l}{mIoU$_n$} & \multicolumn{1}{>{\color{gray}}l}{mIoU$_b$} & \multicolumn{1}{>{\color{gray}}l|}{mIoU$_o$} & \multicolumn{1}{l}{mIoU$_n$} & \multicolumn{1}{>{\color{gray}}l}{mIoU$_b$} & \multicolumn{1}{>{\color{gray}}l|}{mIoU$_o$} & \multicolumn{1}{l}{mIoU$_n$} & \multicolumn{1}{>{\color{gray}}l}{mIoU$_b$} & \multicolumn{1}{>{\color{gray}}l|}{mIoU$_o$} \\ \hline
MaskCLIP~\cite{zhou2022extract}                                      & 12.3                       & 37.2                       & 32.4                        & 16.8                       & 36.6                       & 32.8                        & 16.4                       & 37.0                       & 33.0                        \\
MaskCLIP~\cite{ding2023maskclip}                                      & 15.4 &	38.5&	34.0        & 20.1 &	37.3	&34.0                        & 18.4	&38.7&	34.8	                     \\
OVSeg~\cite{liang2023open}                                          & 17.4                       & 40.0                       & 35.6                        & 25.1                       & 38.7                       & 36.0                        & 22.2                       & 40.2                       & 36.7                        \\
SimSeg~\cite{xu2022simple}                                       & 21.7                       & 37.5                       & 34.4                        & 26.2                       & 37.7                       & 35.5                        & 22.4                       & 37.2                       & 34.3                        \\
FreeSeg~\cite{qin2023freeseg}                                      & 22.1	&35.4	&32.8&		31.6	&29.6&	30.0		&27.2&	29.9&	29.4                 \\

SAN~\cite{xu2023san}  & 25.6                     & 39.5 & 36.8                        & 32.9                       & 39.1                       & 37.9                       & 27.6                       & 42.7                      & 39.8                     \\ \hline
OVFoodSeg$^*$                                    & 28.7\footnotesize{$\pm 0.8$}                       & 44.2\footnotesize{$\pm 0.8$}                     & 41.2\footnotesize{$\pm 0.8$}                       & 37.6\footnotesize{$\pm 1.6$}                     & 43.1\footnotesize{$\pm 0.8$}                       & 42.1\footnotesize{$\pm 0.9$}                       & 31.7\footnotesize{$\pm 1.7$}                      & 44.5\footnotesize{$\pm 1.3$}                       & 42.0\footnotesize{$\pm 1.3$}                         \\
OVFoodSeg                                     & \textbf{30.0}\footnotesize{$\pm 1.2$}                       & 45.5       \footnotesize{$\pm 0.9$}                  & 42.5\footnotesize{$\pm 1.0$}                         & \textbf{38.1}\footnotesize{$\pm 1.0$}                        & 43.0\footnotesize{$\pm 0.8$}                   & 42.0\footnotesize{$\pm 0.8$}     & \textbf{32.8}\footnotesize{$\pm 1.1$}                        & 45.8\footnotesize{$\pm 0.7$}                  & 43.3\footnotesize{$\pm 0.8$}                         \\ \hline
\end{tabular}
}
\caption{The table presents a comparison between existing open-vocabulary segmentation baselines and the proposed OVFoodSeg on FoodSeg103. 
All models are trained on the FoodSeg103 training set and evaluated on the FoodSeg103 test set. OVFoodSeg$^*$ denotes the model based on the FoodLearner without Recipe-1M+ pre-training.}
\label{tab:ap-103-split}
\end{table*}

\begin{table*}[htp]
\resizebox{1.0\textwidth}{!}{
\begin{tabular}{|l|l>{\color{gray}}l>{\color{gray}}l|l>{\color{gray}}l>{\color{gray}}l|l>{\color{gray}}l>{\color{gray}}l|}
\hline
\multicolumn{1}{|l|}{\multirow{2}{*}{Method}} & \multicolumn{3}{c|}{Split 1}                                                          & \multicolumn{3}{c|}{Split 2}                                                          & \multicolumn{3}{c|}{Split 3}                                                          \\
\multicolumn{1}{|c|}{}                        & \multicolumn{1}{l}{mIoU$_n$} & \multicolumn{1}{>{\color{gray}}l}{mIoU$_b$} & \multicolumn{1}{>{\color{gray}}l|}{mIoU$_o$} & \multicolumn{1}{l}{mIoU$_n$} & \multicolumn{1}{>{\color{gray}}l}{mIoU$_b$} & \multicolumn{1}{>{\color{gray}}l|}{mIoU$_o$} & \multicolumn{1}{l}{mIoU$_n$} & \multicolumn{1}{>{\color{gray}}l}{mIoU$_b$} & \multicolumn{1}{>{\color{gray}}l|}{mIoU$_o$} \\ \hline
MaskCLIP~\cite{zhou2022extract}                           &          11.1	&30.4	&26.4                        & 11.0&	26.8	&23.7                       & 9.8&	26.2&	22.8                \\
MaskCLIP~\cite{ding2023maskclip}                           &          12.8	&30.6	&26.9                        & 12.2&	27.0	&24.0                       & 10.5&	27.2&	23.8                \\

OVSeg~\cite{liang2023open}                                          & 15.7	&24.3&	22.5                      &14.4&	21.6&	20.1                        & 13.6&	21.9	&20.2                 \\
SimSeg~\cite{xu2022simple}                                       & 17.5	&35.1	&31.5                     & 14.6&	33.7&	29.8                        & 14.7&	32.4&	28.8                     \\
FreeSeg~\cite{qin2023freeseg}                           &          16.8	&25.1	&23.4                        & 15.9&	22.4	&21.1                       & 15.1&	26.5&	24.2                \\

SAN~\cite{xu2023san}                                     & 20.0&	26.0&	24.8                 & 18.0&	27.3	&25.4                        & 16.6&	26.5	&24.3          \\ \hline
OVFoodSeg*                                    & 22.8\footnotesize{$\pm 0.3$}	&29.9\footnotesize{$\pm 0.2$}&	28.4\footnotesize{$\pm 0.2$}         & 20.5\footnotesize{$\pm 0.2$}&	30.4\footnotesize{$\pm 0.1$}&	28.3\footnotesize{$\pm 0.1$}         & 18.4\footnotesize{$\pm 0.4$}	&28.7\footnotesize{$\pm 0.1$}&	26.6\footnotesize{$\pm 0.2$}                        \\
OVFoodSeg                                     & \textbf{24.0}\footnotesize{$\pm 0.5$}                     & 29.1\footnotesize{$\pm 0.3$}                       & 28.1\footnotesize{$\pm 0.3$}                        & \textbf{21.4}\footnotesize{$\pm 0.4$}                    & 29.8\footnotesize{$\pm 0.3$}                       & 28.1\footnotesize{$\pm 0.3$}                        & \textbf{19.5}\footnotesize{$\pm 0.4$}                      & 28.0\footnotesize{$\pm 0.1$}                       & 26.3\footnotesize{$\pm 0.2$}                       \\ \hline
\end{tabular}
}
\caption{The table presents a comparison between existing open-vocabulary segmentation baselines and the proposed OVFoodSeg on FoodSeg195. All models are trained on the FoodSeg195 training set and evaluated on the FoodSeg195 test set. OVFoodSeg$^*$ denotes the model based on the FoodLearner without Recipe-1M+ pre-training.}
\label{tab:ap-195-split}
\end{table*}

\noindent{\textbf{Implementation}}: We employ the CLIP ViT-L/14~\cite{radford2021learning,dosovitskiy2021an} model as VLM, and initialize FoodLearner with weights of qformer pre-trained via BLIP2~\cite{li2023blip}. In stage I, we initialize $Q$ query tokens ($Q$ set 32 in this paper), each with a channel dimension of 768 in stage I. In stage II, the input image resolution is set to 640$\times$640 to generate visual embeddings. For SAN's classification and mask prediction branches, we set the input image resolution as 640$\times$640 and 320$\times$320. 
We detail the specific hyperparameters for the newly proposed module in the Appendix. For the remaining implementation details, we adhere to the configurations established in SAN~\cite{xu2023san}.

\noindent{\textbf{Metric}}: 
We adopt widely used mean Intersection over Union (mIoU) scores as the metric for segmentation task and 
further use 
mIoU$_n$, mIoU$_b$, and mIoU$_o$ to represent the mIoU for novel classes, base classes, and all classes, respectively.

\subsection{Main Results on FoodSeg103 and FoodSeg195}
Results on the FoodSeg103 and FoodSeg195 datasets are detailed in Table~\ref{tab:ap-103-split} and Table~\ref{tab:ap-195-split}. We train segmentation models and evaluate their performance on FoodSeg103 and FoodSeg195 using multiple class splits to mitigate the impact of randomness on the experiments. For each split, we randomly generate novel classes following the strategy outlined in Section~\ref{sec:exp}. Additionally, we introduce a variant of OVFoodSeg based on FoodLearner without pre-training on the Recipe-1M+ dataset, denoted as OVFoodSeg$^*$.

For both the FoodSeg103 and FoodSeg195 datasets, OVFoodSeg demonstrates a significant improvement over the baseline models for novel classes. For instance, it shows an average improvement of 4.9\% on FoodSeg103 and 3.5\% on the more challenging FoodSeg195, across multiple class splits, compared to the SOTA (State-Of-The-Art) method, SAN.
This strongly validates the effectiveness of the proposed OVFoodSeg framework. It is noted that SimSeg achieves better performance in overall class metrics on FoodSeg195. We attribute this to SimSeg's reliance on a heavily trained mask generator~\cite{cheng2021per}, which offers robustness in complex food scenarios with precise annotations. However, SimSeg is markedly slower than OVFoodSeg, almost 30 times (0.3fps vs 9.8fps on V-100), and underperforms significantly in the crucial novel class metric (6\% in average). Furthermore, the OVFoodSeg$^*$ variant, even without Recipe-1M+ pre-training, also achieves notable improvements of 4.0\% and 2.4\% in novel classes compared with SAN, underscoring the benefits of the image-informed textual representation mechanism.

\subsection{Analysis}
\noindent{\textbf{Objective Losses in Stage I: }}~\label{sec:ap-stageI}In this section, we investigate the impacts of three loss functions (ITC loss, ITM loss, and LM loss) employed in the Stage I pre-training of FoodLearner. In our experiments, we employ different combinations of these losses to pre-train FoodLearner, which is initialized using BLIP2~\cite{li2023blip}. Our segmentation model coupled with the different pre-trained FoodLearner models are trained and evaluated based on the FoodSeg103 dataset. As shown in Table~\ref{tab:ap-stageI}, ITC loss and ITM loss are crucial for achieving competitive results, as they align the representations of food images and text information. Leaving out either ITC or ITM loss leads to a significant deterioration in segmentation performance by 4 to 5\%.  We observe that LM loss enhances the segmentation results by encouraging the query tokens to learn image knowledge related to the corresponding text input.
\begin{table}[htp]
\centering
\resizebox{0.46\textwidth}{!}{
\begin{tabular}{|ccc|c|>{\color{gray}}c|>{\color{gray}}c|}
\hline
ITC & ITM& LM &  mIoU$_n$        & mIoU$_b$        & mIoU$_o$ \\
\hline
   &      \checkmark     &     \checkmark      &   26.5	&43.1&	39.9  \\
      \checkmark     &        &     \checkmark      & 27.3        & 44.2        &  40.9 \\
    \checkmark     &     \checkmark  && 30.8	&45.5	&42.9   \\
    \checkmark & \checkmark     &     \checkmark      & 31.1& 45.3  & 42.5  \\
    \hline
\end{tabular}}
\caption{The impacts of different objective losses used in Stage I FoodLearner Pre-training. The learned FoodLearner of each setting is then used for segmentation learning. We show the results on FoodSeg103.}
\label{tab:ap-stageI}
\end{table}

\begin{figure*}[htp]
\begin{center}
\includegraphics[width=0.95\linewidth]{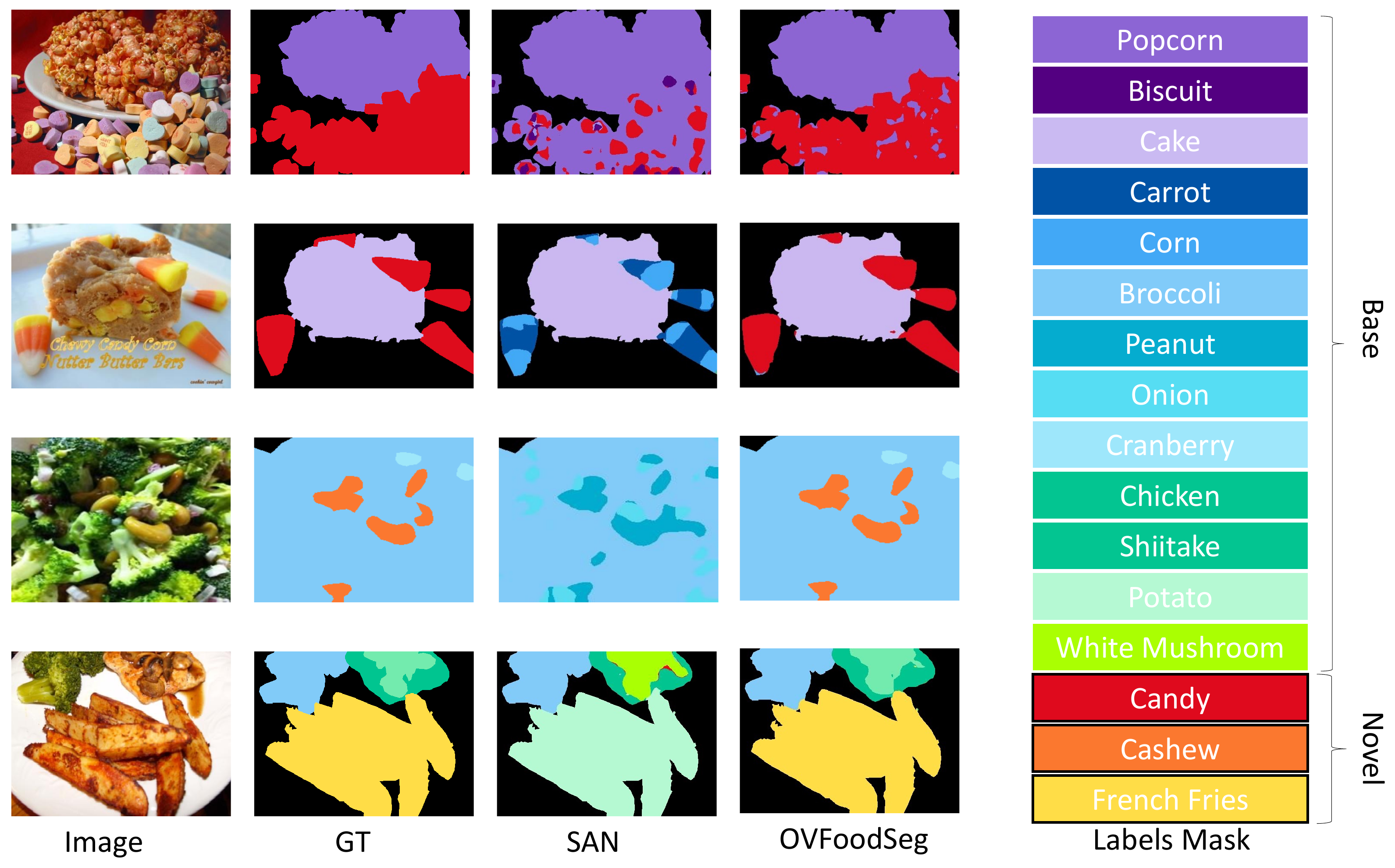}
\end{center}
\vspace{-0.5cm}
\caption{Visulization Results on FoodSeg103 where GT means ground-truth. OVFoodSeg achieves better performance especially for novel classes. 
}
\label{fig:vis}
\end{figure*}

\begin{table}[]
\centering
\resizebox{0.48\textwidth}{!}{
\begin{tabular}{|l|c|>{\color{gray}}c|>{\color{gray}}c|}
\hline
Prompt         & mIoU$_n$                    & mIoU$_b$                    & mIoU$_o$                    \\ \hline
  None    & 30.2   & 44.3    & 41.6                      \\
Default  & 31.1   & 45.3    & 42.5                      \\ 
ViLD~\cite{gu2022openvocabulary}   & 29.4 & 46.7 & 43.3 \\ 
ImageNet~\cite{radford2021learning}    & 29.3 & 48.2 & 44.5 \\ \hline
\end{tabular}}
\caption{Performance comparison of OVFoodSeg using different prompt engineering strategies on FoodSeg103. ``Default'' denotes the single prompt used in the paper.}
\label{tab:103-ab-prompt}
\vspace{-0.45cm}
\end{table}
\noindent{\textbf{Prompt Engineering: }}
Prompt engineering is effective for open-vocabulary segmentation and detection~\cite{gu2022openvocabulary,xu2023san,kuo2023openvocabulary}. This section examines various prompt strategies, employing multiple templates to prompt ingredient names, with each template's image-informed text embedding subjected to average pooling, as shown in Table \ref{tab:103-ab-prompt}. 
Surprisingly, we find that using a default single prompt template, \ie, ``\verb|a photo of {}|'', or even forgoing the use of templates altogether results in significantly better performance in novel classes compared to existing prompt engineering strategies. Existing prompt engineering methods, which average embeddings across templates, may not fully exploit the representation capability of text embeddings. Exploring effective ways to combine multiple template text embeddings is a potential direction for future research.

\noindent{\textbf{Qualitative Results:}}
In Figure~\ref{fig:vis}, we provide qualitative comparisons between SAN and OVFoodSeg on the FoodSeg103 test set.
The results clearly illustrate that OVFoodSeg significantly outperforms SAN, particularly in segmenting novel classes. 
Specifically, the first two rows highlight OVFoodSeg's ability to accurately segment candy with varying appearances, a task at which the baseline model, SAN, fails. 
This effectively demonstrates the superior performance of the proposed OVFoodSeg model in addressing the issue of large intra-class variance.

\section{Conclusion}
In this study, we introduce OVFoodSeg, an innovative open-vocabulary segmentation framework specifically designed for food images. Leveraging the CLIP model, excels in enriching text embeddings with image-specific information, enabled by the innovative FoodLearner and Image-Informed Text Encoder modules. Demonstrating its efficacy on the FoodSeg103 and FoodSeg195 datasets, OVFoodSeg surpasses existing baselines, especially in segmenting novel classes and addressing the substantial intra-class variance prevalent in food imagery. The resultant improvement in segmentation accuracy establishes OVFoodSeg as a new benchmark in the field, paving the way for future advancements in open-vocabulary food image segmentation task.

\section{Acknowledgement}
This research / project is supported by the Ministry of Education, Singapore, under its Academic Research Fund Tier 2 (Proposal ID: T2EP20222-0046). Any opinions, findings and conclusions or recommendations expressed in this material are those of the author(s) and do not reflect the views of the Ministry of Education, Singapore.

{
    \small
    \bibliographystyle{ieeenat_fullname}
    \bibliography{main}
}

\clearpage
\setcounter{page}{1}
\maketitlesupplementary

The implementation details of Stage I and Stage II training are presented in Section \ref{sec:ap-id}. We analyze the failure cases of OVFoodSeg in Section~\ref{app-fail}, and explore the full class training results in Section~\ref{sec:ap-full}. Finally the set of novel classes of these class splits are listed in Section \ref{sec:ap-names}. 

\section{Implementation Details}~\label{sec:ap-id}
{\textbf{Stage I}}: 
We employ the CLIP ViT-L/14 model~\cite{radford2021learning,dosovitskiy2021an} as the CLIP image encoder, and initialize FoodLearner with weights of qformer pre-trained via BLIP2~\cite{li2023blip}. We initialize $Q$ query tokens ($Q$ set 32 in this paper), each with a channel dimension of 768.
The model undergoes training for 10 epochs by AdamW optimizer, processing images at a resolution of 224$\times$224 pixels within batches of 100. We utilize cosine learning rate schedules, starting with an initial learning rate of 1e-4 and decaying to an ending rate of 1e-5, coupled with a weight decay set at 0.05. A linear warmup strategy is applied during the initial 5000 iterations, starting at a learning rate of 1e-6. 

\noindent{\textbf{Stage II}}: 
We utilize the CLIP ViT-L/14 model for both image and text encoding, initializing the FoodLearner and query tokens with weights pre-trained from Stage I. 
Ingredient class names are prompted with the template ``\verb|A Photo of {}|'' to produce text tokens. 
The input image resolution is set to 640$\times$640 to generate visual embeddings. For SAN's classification and mask prediction branches, we set the input image resolution as 640$\times$640 and 320$\times$320 respectively. For FoodSeg103, the model undergoes training for 10,000 iterations using the AdamW optimizer, with a batch size of 8. A Poly learning rate schedule is employed, beginning with an initial learning rate of 1e-4, coupled with a weight decay set at 0.0001. For FoodSeg195, the model is trained over 20,000 iterations, maintaining consistency with the other settings used for FoodSeg103. For the remaining implementation details, we adhere to the configurations established in SAN~\cite{xu2023san}.
\section{Failure Case Analysis}~\label{app-fail}
\begin{figure}[htp]
\begin{center}
\includegraphics[width=0.98\linewidth]{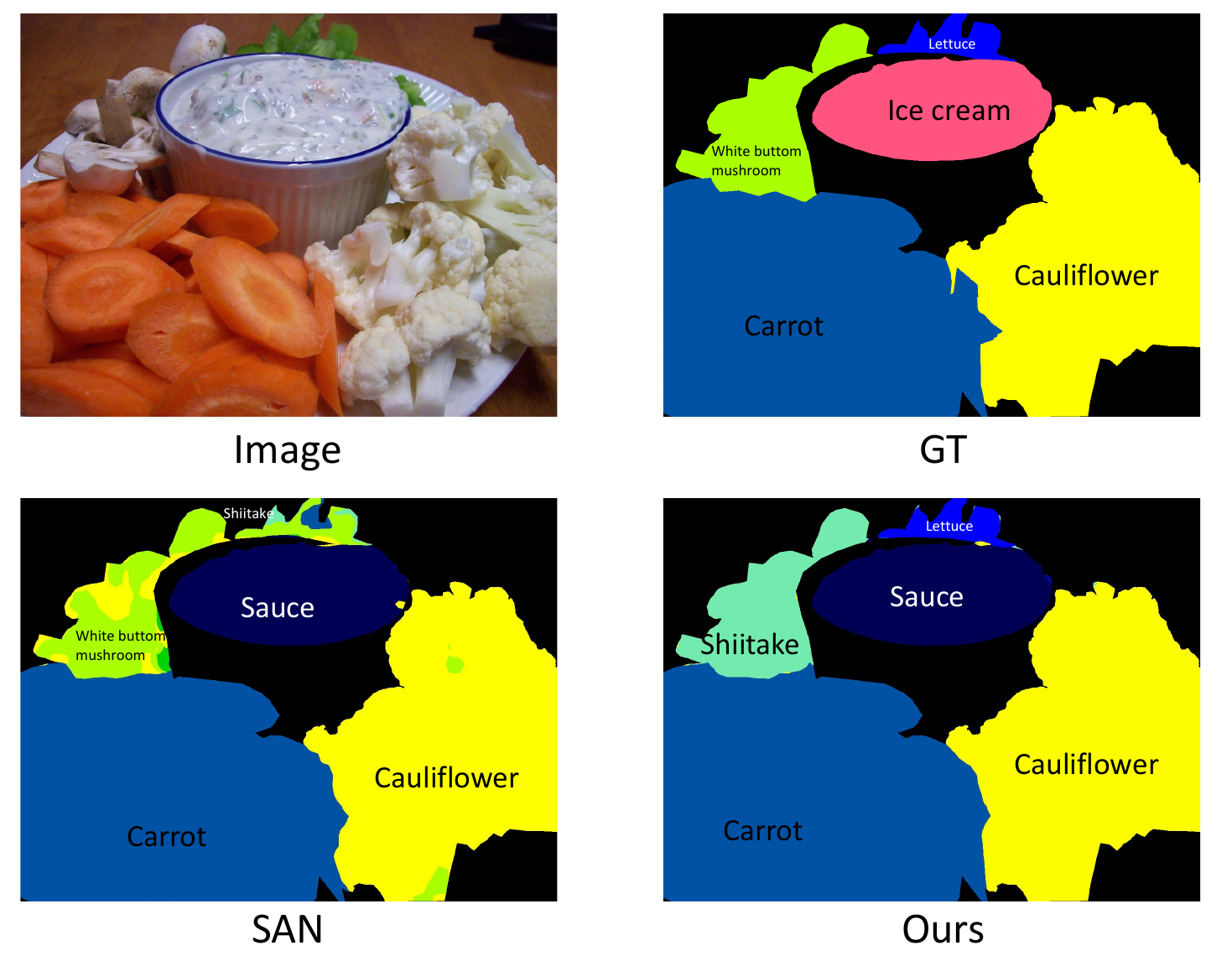}
\end{center}
\vspace{-0.5cm}
\caption{Failure cases of OVFoodSeg on FoodSeg103 (Split 1) where GT means ground-truth. In this example, OVFoodSeg incorrectly classified ``white button mushroom'', a novel class, as ``shiitake'', which is a base class.
}
\label{fig:app-fail}
\end{figure}

In this section, we analyze the failure cases of OVFoodSeg, particularly focusing on the classes that perform worse than the baseline SAN model. Among the 20 novel classes of FoodSeg103 Split 1, 3 classes perform worse than the baseline. Notably, the \verb|white button mushroom| class shows the most significant underperformance, with its results being approximately 8.9\% lower than the baseline (2.2\% vs 11.1\%). We visualize the prediction results of OVFoodSeg in Figure~\ref{fig:app-fail}. The figure illustrates that OVFoodSeg mistakenly classifies the novel class \verb|white button mushroom| into the base class  \verb|shiitake|, which bears a high visual resemblance to the target novel class. 
The confusion between novel and base classes poses a significant challenge in the open-vocabulary setting, and tackling this issue is a key focus for our future research.

\section{Full Class Training}~\label{sec:ap-full}
In this section, we compare OVFoodSeg and SAN trained with both novel and base classes of FoodSeg103. For a fair comparison, we use FoodLearner without Stage I pretraining (OVFoodSeg*). From Table \ref{tab:full-class}, OVFoodSeg* significantly outperforms SAN in the full class training mode, further proving the effectiveness of our proposed image-informed textual representation mechanism.

\begin{table}[htp]
\centering
\resizebox{0.35\textwidth}{!}{
\begin{tabular}{|l|c|c|c|}
\hline
         & mIoU                    & mAcc                    & pAcc               \\ \hline
  SAN    & 41.6   & 56.2    & 67.1                      \\
OVFoodSeg*  & 45.4   & 60.4    & 71.0                      \\ 
\hline
\end{tabular}}
\caption{Performance comparison of OVFoodSeg* and SAN trained with both novel and base classes on FoodSeg103. Here, mIoU, mAcc and pAcc denote mean IoU, mean accuracy and pixel-wise accuracy respectively.}
\label{tab:full-class}
\end{table}
\section{Novel Classes in Multiple Class Splits}~\label{sec:ap-names}
In this section, we detail the novel ingredient classes utilized in each class split.
\subsection{FoodSeg103}
\noindent{\textbf{Split 1:}}
\begin{verbatim}
    candy, french fries, ice cream, wine, 
    coffee, cashew, pineapple, sausage, 
    lamb, crab, pie, seaweed, lettuce, 
    pumpkin, bamboo shoots, celery stick, 
    cilantro mint, cabbage, bean sprouts, 
    white button mushroom
\end{verbatim}

\noindent{\textbf{Split 2:}}
\begin{verbatim}
    egg tart, coffee, date, blueberry, 
    raspberry, kiwi, chicken or duck, 
    soup, bread, hanamaki baozi, 
    eggplant, kelp, seaweed, ginger, 
    carrot, asparagus, cabbage, onion, 
    green beans, salad
\end{verbatim}

\noindent{\textbf{Split 3:}}
\begin{verbatim}
    french fries, cake, juice, 
    red beans, apricot, raspberry, 
    melon, watermelon, shellfish, 
    shrimp, kelp, seaweed, 
    spring onion, okra, carrot,
    cilantro mint, snow peas, 
    king oyster mushroom, shiitake,
    white button mushroom
\end{verbatim}

\subsection{FoodSeg195}
\noindent{\textbf{Split 1:}}
\begin{verbatim}
    popcorn, cheese, cake, milk, date, 
    avocado, raspberries, lemon, 
    pineapple, grape, melon, steak, 
    pork, sausage, bread, pizza, pasta, 
    rice, garlic, kelp, broccoli,
    celerystick, cilantro mint, 
    pork belly, edamame, ketchup, 
    fish cake, fish balls, rice cake, 
    lotus root, daylily, durian, 
    thosai, tangyuen, idli, spaghetti, 
    nai bai, kangkong, yam, beef
\end{verbatim}

\noindent{\textbf{Split 2:}}
\begin{verbatim}
    blueberry, melon, steak, shrimp,
    baozi, pasta, noodle, pie, tomato, 
    ginger, okra, lettuce, others, 
    wolfberry, pig blood curd,
    meat skewer, meatballs, edamame,
    curry sauce, salad sauce,
    garlic sauce, porridge, amaranth,
    honey dews, papaya, beehoon,
    lasagna, macaroni, puri, oyster,
    celery, sweet potato, beef
\end{verbatim}

\noindent{\textbf{Split 3:}}
\begin{verbatim}
    kiwi, orange, fried meat, baozi,
    garlic, spring onion, ginger,
    lettuce, white radish, asparagus,
    bamboo shoots, bean sprouts,
    oyster mushroom, beef ribs,
    minced beef, pork belly,
    pork skin, pork liver,
    shredded pork, meat skewer,
    edamame, barbecued pork sauce,
    fish tofu, fried banana leaves,
    bitter gourd, agaric, burger buns,
    guava, mochi, bun, mussel, celery,
    beef
\end{verbatim}

\end{document}